%% file: main1.tex
\pdfoutput=1
\documentclass[11pt]{article}
\usepackage[]{EACL2023}

\usepackage{times}
\usepackage{latexsym}
\usepackage{graphicx}

\usepackage[T1]{fontenc}

\usepackage[utf8]{inputenc}

\usepackage{microtype}

\usepackage{textcomp}
\usepackage{amsmath}
\usepackage[varg]{newtxmath}

\usepackage{multirow}

\usepackage{adjustbox}
\usepackage{xcolor}
\usepackage{pgfplots}
\pgfplotsset{compat=1.16}

\definecolor{forestgreen}{rgb}{0.13, 0.55, 0.13}

\usepackage[scaled=0.8300]{beramono}
\usepackage{linguex}
\title{Spelling convention sensitivity in neural language models}

\author{Elizabeth Nielsen$^\dag$ \enskip Christo Kirov$^\circ$ \enskip Brian Roark$^\circ$\\
  $^\dag$School of Informatics, University of Edinburgh, UK \hspace*{0.2in} $^\circ$Google\\
  \texttt{e.nielsen@ed.ac.uk \hspace*{0.2in}\{ckirov,roark\}@google.com}}

\begin{document}
\maketitle

\input{abstract_acl2023}

\input{introduction1}
\input{background1}
\input{datamodels}
\input{exp0corpus}

\input{exp1score}
\input{gpt2}

\input{exp2corpus}

\input{conclusion1}
\input{ack}
\input{limitations}
\input{ethics}


\bibliographystyle{acl_natbib}
\bibliography{anthology,main}

\appendix

\input{appendix}

\end{document}

%% file: abstract_acl2023.tex
\begin{abstract}
We examine whether large neural language models, trained on very large collections of varied English text, learn the potentially long-distance dependency of British versus American spelling conventions, i.e., whether spelling is consistently one or the other within model-generated strings. 
In contrast to long-distance dependencies in non-surface underlying structure (e.g., syntax), spelling consistency is easier to measure both in LMs and the text corpora used to train them, which can provide additional insight into certain observed model behaviors.
Using a set of probe words unique to either British or American English, we first establish that training corpora exhibit substantial (though not total) consistency. A large T5 language model does appear to internalize this consistency, though only with respect to observed lexical items (not nonce words with British/American spelling patterns).
We further experiment with correcting for biases in the training data by fine-tuning T5 on synthetic data that has been debiased, and find that finetuned T5 remains only somewhat sensitive to spelling consistency. Further experiments show GPT2 to be similarly limited.
\end{abstract}

%% file: introduction1.tex
\section{Introduction}

The probabilities that neural language models (LMs) assign to strings can be used to assess how effectively they capture linguistic dependencies found in their training data.
Much as in psycholinguistic experiments on human language speakers, we can present LMs with strings both with and without agreement in key dependencies and measure the assigned probabilities to determine whether the model has learned these linguistic generalizations or not (see e.g., \citealt{futrell2018rnns}).  For example, sentences both with and without subject/verb number agreement (but otherwise identical) can be used to assess whether the model accounts for that particular dependency, even over long distances.
Various long-distance dependencies have been investigated in this manner, from purely linguistic phenomena such as syntactic dependencies (e.g., \citealt{gulordava-etal-2018-colorless}) to extra-linguistic phenomena such as socio-cultural biases (e.g., \citealt{rudinger-etal-2018-gender}). 

In this paper, we examine dependencies based on orthographic cues to language variety.  Many LMs are trained on large corpora scraped from the web, and data from different language varieties are often combined.
For example, LMs trained on web-scraped English (e.g., the WebText Corpus of \citealt{radford2019language}) encounter British English, North American English, and multiple World Englishes.
Likewise, Spanish web corpora may include several distinct varieties of Latin American Spanish, as well as Iberian Spanish (e.g., \citealt{KILGARRIFF201312}).  Here we use differences between British and American English spelling conventions to ask whether LMs trained on large and diverse collections of English learn to apply these conventions consistently within the same span of text.
For example, if the British spelling of the word {\it labour\/} appears in a sentence prefix, will the LM assign higher probabilities to continuations that maintain British spelling conventions (e.g., {\it organisation\/}) over those that have American-spelled forms ({\it organization\/})? To the extent that such models are used within response generation systems or for next word prediction in virtual keyboards, maintaining such consistency would be strongly desirable so users receive results appropriate for their locale. 

Of course, as with any such dependencies, models can only learn generalizations that are present in the data, so we also look at the degree to which corpora used to train the large LMs (LLMs) that we investigate (as well as a few others) demonstrate spelling convention consistency. 
Assessing whether syntactic or semantic generalizations are learned by models trained on noisy, errorful and inconsistent data is complicated by the difficulty in quantifying the actual degree of consistency of the dependency in the data itself.
In contrast to structural linguistic generalizations or other implicit information, the explicitness of spelling conventions permits straightforward corpus analysis in addition to model probing, providing another avenue for explaining model performance.

The results of our data analysis are presented in \S\ref{sec:corpus}.  We find that relevant web-scraped English text used to train LLMs unsurprisingly does not provide perfect consistency\,---\,and further that it is heavily skewed towards American spelling conventions\,---\,but that it provides as much or more consistency than some curated corpora such as the British National Corpus \cite{BNC2007}.  We then present methods, in \S\ref{sec:score}, to measure the degree to which two neural LLMs -- T5 \cite{raffel-et-al-2020} (both with and without additional finetuning) and GPT2 -- exhibit spelling variation consistency.  We find that T5 without finetuning demonstrates a general preference for consistency, but that this preference is weaker for British than American English and does not extend robustly to nonce words.  Finetuning T5 on a synthetically modified portion of the British National Corpus reduces the preference for American English. We then modify our conditional probability calculations to allow demonstration of similar patterns of model behavior for GPT2, a very differently architectured and trained LLM~ \cite{radford2019language}.
Lastly, in \S\ref{sec:corpuspartdeux}, we take a slightly deeper dive into the kinds of (and reasons for) spelling convention inconsistencies in some corpora analyzed in \S\ref{sec:corpus}.

Overall, we demonstrate that, while T5 and GPT2 display some sensitivity to spelling convention differences, this cannot be relied on to produce consistent generated output. If reliable spelling consistency is an application requirement, additional post-processing may need to be applied to LLM output.



This paper makes several key contributions.  First, we provide methods for straightforwardly assessing the ability of LLMs to capture certain well-attested long-distance dependencies in English, and demonstrate the strengths and shortcomings of two well-known models in doing so. This opens up the possibility of exploratory studies in languages where such conventions are less well documented.  In contrast to the most heavily investigated types of long-distance dependencies (e.g., syntactic), the (previously unexplored) dependency of spelling convention consistency is directly observable in the surface string and hence is relatively easy to assess in both models and data.  As a result, it can be seen as a useful task for assessing LM learning in general.
We also document the degree to which web-scraped corpora exhibit spelling consistency, making clear that the models have plenty of room for improvement. However, American English is shown to be far more heavily represented in the training corpora than British English, to the point that performance for British English is demonstrably far worse than for American English, something that language generation or word prediction systems must address for equitable performance.

%% file: background1.tex
\section{Background}\label{sec:background}

\subsection{Dependencies and LMs}

Much of the work investigating whether large language models capture long-distance linguistic generalizations has focused on non-surface dependencies, such as co-reference.
In order to correctly identify that two expressions refer to the same entity, models often need to identify complex syntactic relationships (e.g., c-command), or build a model of entities over an entire discourse (e.g., \citealt{clark2016deep}).
Despite this complexity, LLMs have shown some promise as general-purpose co-reference resolvers \cite{joshi-etal-2019-bert}.
This suggests that they can learn to model complex long-distance dependencies.

Other research has shown more directly that LLMs model syntactic dependencies. 
A common methodology is to compare an LM's surprisal directly to psycholinguistic data \cite{futrell2018rnns}. 
If the LM still performs like a human on examples that require modeling hierarchical relationships between tokens, this suggests that the LM has learned some part of the more complex syntactic structure of the language.
Work such as \citet{futrell2018rnns} has shown that a recurrent neural network language model achieves surprisal rates that mimic human processing, including in these syntactically complex situations.
This suggests that an RNN LM can be sensitive to complex syntactic relationships as well.
Similar methods have been used to show LMs learning syntactic dependencies in \citet{linzen-etal-2016-assessing}, \citet{frank-etal-2016}, and \citet{brennan-2020}.

Another class of methods for assessing whether LMs learn complex syntactic dependencies involves probing the models themselves to evaluate whether syntax-like relationships between tokens can be discovered.
Details of their methods vary widely, but \citet{clark-etal-2019-bert-DUP}, \citet{hewitt-manning-2019-structural}, and \citet{lin-etal-2019-open} all suggest that many LMs learn complex syntactic dependencies.

In contrast, the topic of the current paper -- spelling convention dependencies -- is a relatively surface-level dependency.
A model does not need to capture the syntactic or semantic relationship between two words in order to evaluate spelling consistency, rather simply their co-occurrence.
Given prior results showing that LMs can and do learn complex semantic and syntactic relationships between words, one might expect that a relatively simple dependency like spelling convention should be easy for an LM to learn.

\subsection{Spelling variation}
As discussed by \citet{berg2017self}, the orthography of English has never been regulated by an official body, but has rather emerged dynamically over time.
Dictionaries played a key role in settling spelling conventions, with Samuel Johnson's \shortcite{johnson1755dictionary} dictionary being the key source of contemporary British spelling conventions and Webster's \shortcite{webster1828dictionary} dictionary the key source of contemporary American spelling.  
The latter included spelling reforms such as using the suffix {\it -or\/} instead of {\it -our\/} for certain words, e.g., {\it labor\/} instead of {\it labour\/}.  
These reforms were adopted in American spelling but not in British spelling conventions.

This history makes English an interesting case study for spelling variation in particular.
Languages that have historically had centralized regulatory institutions, such as the French or Royal Spanish Academies, have much less purely orthographic variation.
For example, despite many lexical differences, there are few spelling differences between Iberian and Latin American Spanish.
On the other hand, there are many language situations that have considerably more spelling variation.
For example, speakers of South Asian languages that are traditionally written with Brahmic or Arabic scripts often write using the Latin alphabet in contexts like SMS messages and social media \cite{roark-etal-2020-processing}.
This kind of informally romanized text presents many spelling variations due to these languages' lack of orthography in the Latin script.
The well-documented nature of English spelling variation and its close ties to standardized regional varieties make it a good initial case study for whether LLMs learn systematic variation in the data.  If so, such models may be useful in more exploratory studies, such as the above-mentioned scenario where no official orthography exists.

As far as we are aware, the issue of spelling convention consistency in language models has not been investigated.
\citet{nguyen-grieve-2020-word} looked at whether word embeddings are {\it robust\/} to spelling variation, not whether generative language models capture spelling consistency.  
That paper focused mainly on the kinds of variation that arise in informal social media text, but they also examined British versus American spelling.  
Unsurprisingly, they found that cosine similarity between British and American spelled variants are high relative to other patterns of informal spelling variability.

\subsection{Prompting LMs}

In the present work, we construct prompts to measure the probability assigned to various tokens by LLMs. 
In constructing these prompts, we take into account the findings of recent work on prompting LMs.
Our work is different from the sort of prompting described by these papers, which generally includes features such as task-specific prefixes containing instructions (e.g., \citealt{raffel-et-al-2020}), verbalized class labels (e.g., \citealt{schick-schutze-2021-exploiting}), or in-context learning (e.g., \citealt{brown-et-al-2020}), none of which are present in our approach.
However, work such as \citet{webson-pavlick-2022-prompt} has shown large effects due to small variations in the wording of prompts, even if the reasons for these effects are not apparent.
Therefore, we choose to present the model with several different prompts and average the probabilities over all prompts, in order to account for possible variation.

%% file: datamodels.tex
\section{Data and models}  \label{sec:datamodels}
To assess the spelling convention consistency of data and models, we use a list of British and American English spelling differences that is part of the open-source American British English Translator.\footnote{\url{https://github.com/hyperreality/American-British-English-Translator}} We used the 1706 word pairs in the {\tt data/american\_spellings.json} file at that site. This list includes American and British spelling variants for words with common differences such as {\it -or\/}/{\it -our\/} (e.g., vapor/vapour), $\mbox{{\it -ize\/}/{\it -ise\/}}$ (realize/realise), consonant doubling (modeling/modelling), {\it -er\/}/{\it -re\/} (liter/litre), along with some number of term-specific spelling differences (aluminum/aluminium).  We use this list to create prompts for probing the language models and to establish the consistency of usage within corpora, i.e., whether strings found in this list consistently follow one convention or the other when they co-occur.

For model probing, we examine T5 \cite{raffel-et-al-2020}, a general purpose encoder-decoder model. We use the t5-large architecture variant on the T5X codebase,\footnote{\url{https://github.com/google-research/t5x/blob/main/docs/models.md\#t5-checkpoints}} which has approximately 770M parameters. For English, T5 is (pre-)trained using a span corruption objective on the Colossal Clean Crawled Corpus (C4), an English language collection derived from Common Crawl \cite{raffel-et-al-2020}.\footnote{\url{http://commoncrawl.org/}} 

We also examine GPT2, for which we use the open-source HuggingFace implementation \cite{radford2019language}. Unlike T5, GPT2 is a purely autoregressive language model rather than an encoder-decoder sequence-to-sequence model. It is trained to perform next-word prediction rather than fill in corrupted spans of text. GPT2 is built on OpenAI's WebText corpus \cite{radford2019language}, of which there is an open-source variant available.\footnote{\url{https://skylion007.github.io/OpenWebTextCorpus/}} 

We examine C4 and OpenAI's WebText corpus for spelling convention consistency, along with several other corpora: English Wikipedia (downloaded 06-21-2020); the Billion Word Benchmark \cite{chelba2013one}, which is a collection of newswire text; and the British National Corpus \cite{BNC2007},\footnote{\url{http://www.natcorp.ox.ac.uk/}} which is a balanced corpus of both written and spoken material.\footnote{Code for querying corpora and generating prompts, as well as other relevant data and code, can be found at \url{https://github.com/google-research/google-research/tree/master/spelling_convention_nlm}.}

%% file: exp0corpus.tex
\section{Training corpora consistency}\label{sec:corpus}

To examine spelling consistency in training data, we made use of the list of spelling variants and the five corpora mentioned in Section \ref{sec:datamodels}: C4, the OpenWebText Corpus (OWT), English Wikipedia (EngWiki), the Billion Word Benchmark (BWB), and the British National Corpus (BNC).  
We convert all strings in each corpus to lowercase, and treat all characters outside of the a--z range as whitespace for tokenization. We look for exact matches of list items in the resulting whitespace-delimited tokens.

Let $V_{\mathrm{US}}$ be the US spelling variants\footnote{For convenience, we use US as shorthand for American and UK as shorthand for British.} of the words in the list and $V_{\mathrm{UK}}$ the UK spelling variants.  
For each corpus $C$, let $s^k=w_1\ldots w_{|s^k|}$ represent the $k$th string in the corpus, consisting of $|s^k|$ words.  We extract all pairs of words $(w_i, w_j)$ from $s^k$ such that $i < j$ and $w_i,w_j \in V_{\mathrm{US}} \bigcup V_{\mathrm{UK}}$. Each extracted pair $(w_i, w_j)$ is placed into one of three classes: the pair is (1) {\it US-matched\/} if $w_i,w_j\in V_{\mathrm{US}}$; (2) {\it UK-matched\/} if $w_i,w_j\in V_{\mathrm{UK}}$; and (3) {\it mismatched\/} otherwise. We then aggregate the counts for pairs in these three bins across all strings in the corpus.

\begin{table}
\centering
\begin{tabular}{l|r|ccc}
& total \# of ~~ & \multicolumn{3}{c}{X-matched \%}\\		
Corpus	& word pairs~~ & US & UK & mis \\\hline
C4 & 542,755,756 & 74.6 & 14.7 & 10.8\\
OWT & 42,255,261 & 79.7 & 11.5 & ~8.8\\
EngWiki & 1,527,529 & 58.0 & 26.5 & 15.4\\
BWB & 442,733 & 67.5 & 23.6 & ~8.9\\
BNC & 74,072 & 14.5 & 64.8 & 20.8\\\hline
\end{tabular}
\caption{\footnotesize Study of word pairs found in the same string from either UK or US spelling list over corpora of different sizes and characteristics, with percent of {\it US\/}-matched, {\it UK\/}-matched and {\it mis\/}matched US/UK pairs.}\label{tab:corpora}
\end{table}

Table \ref{tab:corpora} presents the number of pairs extracted from each corpus and the percentage of those within each class.  Several things jump out from these results.  First, all of the corpora, other than the British National Corpus, have significantly more US-matched pairs than UK-matched pairs, with OWT and C4 being the most skewed towards US-matched pairs.  This likely indicates a heavy overall skew towards US spelling variants, leading to a high prior probability of US spelling variants in LLMs. Second, the percentage of extracted pairs that are mismatched are non-negligible, however there is a lot of consistency.  For example, in the C4 corpus, if a word from $V_{\mathrm{UK}}$ is the first word of a pair, the probability that the next word will also be from $V_{\mathrm{UK}}$ is nearly three times the probability that it is from $V_{\mathrm{US}}$.\footnote{Mismatched pairs in all corpora are roughly equally split between having $V_{\mathrm{US}}$ or $V_{\mathrm{UK}}$ words first. Hence, for C4, 5.4\% of pairs are $V_{\mathrm{UK}}$ followed by $V_{\mathrm{US}}$ words (half of the mismatched probability), while 14.7\% are $V_{\mathrm{UK}}$ followed by $V_{\mathrm{UK}}$.}  Finally, both English Wikipedia and the British National Corpus have somewhat elevated levels of mismatch compared to the other corpora, something we look at more closely in Section \ref{sec:corpuspartdeux}.

Having established that the level of mismatch in the C4 corpus used to train T5 is at the lower end observed in the data we examined,\footnote{We note again the benefit of these explicit surface-level dependencies -- we can easily assess the prevalence/consistency of the training data, in contrast to structural dependencies.} we now move on to examine whether the trained models pick up on these dependencies.

%% file: exp1score.tex
\section{Language model consistency} \label{sec:score}



From the dictionary presented in Section~\ref{sec:datamodels}, we kept only the words that can be described by a small number of rules, e.g., the variation between \textit{-ize} and \textit{-ise}, etc, leaving us with 1266 options.
For efficiency, we sample $\approx$16k prompt-target pairs (16028) from all possible $1266^2$ combinations.


To eliminate all sources of variation besides the pair of words being tested, we created several template sentences into which we can insert pairs of words. The full set of templates is presented in Table~\ref{tab:PromptSet} in Appendix \ref{sec:appendixprompt}.
Several considerations informed how we formulated the templates so that they work for all the tokens we wanted to test. 

First and most obviously, we need to ensure that all tokens in a template are variety-neutral.
This ensures that the probability of any of our test words being British or American will not be swayed by any regional bias in the template. While neutrality is difficult to enforce perfectly within a single frame, we hope that by using multiple different templates, we can mitigate unknown sources of bias via averaging. 

Second, we need templates that will be syntactically and semantically acceptable, regardless of the inserted tokens.
LLMs may assign low probability to tokens that result in grammatically unacceptable or semantically unlikely sentences, and we want to avoid introducing this source of variation.
This is challenging, since the tokens we are testing include different parts of speech and come from very different semantic domains, hence there are few contexts where all tokens would be acceptable.

\begin{table*}[t]
    \centering
    \begin{tabular}{l|c|cc||cc||cc}
    \hline
    && \multicolumn{2}{c||}{T5} & \multicolumn{2}{c||}{T5+FT} & \multicolumn{2}{c}{C4}\\
    Condition & Word 1 & \multicolumn{2}{c||}{Word 2} & \multicolumn{2}{c||}{Word 2} & \multicolumn{2}{c}{Word 2}\\ 
    & & US & UK & US & UK & US & UK\\
    \hline
    Adjacent &
     US & 0.86 & 0.14 & 0.66 & 0.34 & 0.91 & 0.09\\
    & UK & 0.39 & 0.61 & 0.44 & 0.56 & 0.38 & 0.62
 \\\hline
         Non-adjacent & 
     US & 0.83 & 0.17 & 0.69 & 0.31 & 0.93 & 0.07\\
    & UK & 0.48 & 0.52 & 0.43 & 0.57 & 0.27 & 0.73\\\hline
    \end{tabular}
    \caption{\footnotesize Conditional probability of Word 2, given a template with Word 1, given by T5 (no finetuning) and T5+FT (finetuned on synthetic balanced BNC data). For each instance, the probability has been normalized over each condition (corresponding to each row for the model).  We also present the conditional probabilities from pairs found in the training corpus C4.}
    \label{tab:cond}
\end{table*}

Fortunately, this problem has an analogue in linguistics: linguists interested in detailed phonetic description often elicit tokens in set contexts to eliminate extraneous sources of acoustic variation \cite{bowern2008}.
The approach these linguists often take is to use a template that \textit{mentions} the tokens in question rather than \textit{using} them.
We follow this approach, and use templates similar to \ref{ex:template1}, which contain a list of word mentions.

\ex. \textit{My preferred words are ..., ..., and tree.} \label{ex:template1}

We then substitute pairs of words from our dictionary into the spaces marked with ellipses, both with consistent and inconsistent spelling conventions. 
In other words, given the pair of dictionary entries \textit{realize\/}/\textit{realise} and \textit{center\/}/\textit{centre}, we use the template above to generate the four test sentences:

\ex. \label{ex:template} \a. US/US: \textit{My preferred words are\\ \textbf{realize}, \textbf{center}, and tree.}
\b. US/UK: \textit{My preferred words are\\ \textbf{realize}, \textbf{centre}, and tree.}
\c. UK/US: \textit{My preferred words are\\ \textbf{realise}, \textbf{center}, and tree.}
\d. UK/UK: \textit{My preferred words are\\ \textbf{realise}, \textbf{centre}, and tree.} 

We use T5 to score the probability of generating the second bolded word, as shown in Example \ref{ex:template}, given the first. 

In the above template, the two words are adjacent in the string.  We also include a non-adjacent condition, which augments the templates by adding ten variety-neutral tokens between the bold-face words. For the above sample, the non-adjacent variant would be:

\ex. \textit{My preferred words are ..., flower, interesting, jump, ponderous, sky, skipping, desk, small, ladder, lovely, ..., and tree.}

Since T5 is a seq2seq model trained on a span-corruption objective, we present a prompt that includes a priming word and a blank span token representing the second word: 
\ex. \hspace*{-0.2in}My preferred words are \textbf{flavour},\\
\hspace*{-0.2in}<{\sc blank-span}-1>, and tree

The decoder then scores an output string that replaces the blank, but represents the known inputs with span markers instead: 

\ex. \hspace*{-0.2in}<{\sc input-span}-1> \textbf{harbour} <{\sc input=span}-2>

Thus we are effectively computing the probability that the blank span will be filled with a particular word (with a US or UK spelling), given the visible input sentence (which contains a US or UK primer)\,---\,P(``harbour'' | ``My preferred words are flavour, ..., and tree.'').


We report a few different measures to give a picture of how strongly each model prefers spelling consistency: mean conditional probabilities, prediction accuracy and mutual information.  We then examine behavior with nonce words.

\subsection{Measure 1: conditional probability tables} \label{sec:cond}

The first measure we use to show the preferences of each model is a 2x2 table of the conditional probability of the second probe word, given the first.
For ease of interpretation, we normalize the conditional probabilities for each conditioning word as though the two alternative second words (US and UK) are the only possibilities, i.e., the two conditional probabilities are made to sum to 1.  That is, $P(UK|US) + P(US|US) = 1$ and $P(US|UK) + P(UK|UK) = 1$ for each example.
These conditional probabilities are then averaged over the whole test corpus (16028 word pairs replicated across 29 template sentences\footnote{For information on the variance across prompts, see Appendix \ref{sec:appendixprompt}.} for a total of 464812 samples) for both the adjacent and non-adjacent conditions.
Table \ref{tab:cond} presents these mean conditional probabilities for base T5 and T5 finetuned (TF+FT) on a synthetic balanced corpus derived from the BNC (see \S\ref{sec:finetuning}), alongside conditional probabilities calculated from the pairs extracted for the analysis in Table \ref{tab:corpora} from their training corpus (C4), under the same adjacent and non-adjacent conditions.\footnote{The conditional probabilities from C4 are simply the probability that Word 2 is from the UK or US class given the class of Word 1, with extracted pairs split by whether the words were adjacent or not in the string.  Adjacent pairs account for roughly 1\% of all pairs in the corpus.}


As can be seen from these results, T5 shows a preference for spelling consistency in both the adjacent and non-adjacent conditions --- probabilities for both the consistent US and consistent UK conditions are higher than the probabilities for the respective inconsistent conditions. 
The differences are notably larger in the adjacent conditions than the non-adjacent conditions, indicating that the preference for spelling consistency attenuates somewhat over longer strings.
The model also shows a preference for US forms overall, assigning a higher probability to a US form after a UK form than to a UK form after a US form.
This is likely due to US forms being over-represented in the training data, leading to high prior probability.




Comparing the model and corpus columns in  Table \ref{tab:cond}, the degree of consistency preference displayed by T5 in the adjacent condition is actually very similar to the consistency levels in the C4 training corpus (similarly replicating the bias for US forms). However, C4 is much more consistent in the non-adjacent condition than T5, indicating that the model is failing to capture some long-distance dependencies.


\subsection{Finetuning on synthetic data} \label{sec:finetuning}
Finding naturally occurring English text using perfectly consistent spelling conventions of sufficient size to help improve a model's consistency may be difficult, given the results presented in Table \ref{tab:corpora}.  It would be useful, however, to determine if T5 could be finetuned with some resource to exhibit better spelling consistency.  To that end, we created a synthetic version of the BNC, which was changed to exhibit perfect consistency of British and American spelling conventions for the words in our lexicon.

This synthetic BNC corpus was produced as follows.  Using our list of spelling variants, we identified strings in the corpus that contained an instance of either the American or British spelling.  We then produced a synthetic consistent American spelling version of these strings by using the American spelling of all of the words, along with a synthetic consistent British spelling version of these strings by using the British spelling of all of the words.  The resulting corpus is thus balanced between American and British spelling for these 1706 words, and every sentence is consistent in using one convention or the other. In total, the synthetic corpus contains 954238 sentences,\footnote{In our testing, this was not enough data to reliably train a T5-large LLM from scratch.} equally split between US and UK spelling conventions. A small random subset of 2560 sentences was reserved for validation, and T5 was finetuned on the rest. Finetuning used the same span-filling masked LM task used for pretraining, with dropout set to 0.1, and the loss normalizing factor set to 233472 as suggested in the T5 documentation. Fine-tuning started at the default T5-large checkpoint, which represents 1000700 steps, and proceeded another 99300 steps at a batch size of 128.

As seen in Table \ref{tab:cond}, finetuning on this synthetic corpus does not appear to improve overall spelling consistency -- quite the opposite. However, it does have at least two interesting effects. First, as might be expected, the overwhelming preference for US English shown by base T5 is reduced. Furthermore, the finetuned model is better able to retain long-distance information --- there is no dropoff in consistency between the adjacent and non-adjacent conditions as seen for T5 without finetuning.

\subsection{Measure 2: prediction accuracy} \label{sec:acc}

While the conditional probabilities in Table \ref{tab:cond} show the overall preferences of the models over the test set, we also want a measure that captures how often the LLMs assign consistent pairs a higher probability than inconsistent pairs.
In Table \ref{tab:acc} we show the percentage of the test set examples for which each model predicted consistency over inconsistency. The results show a similar pattern as the conditional probability measures in Table \ref{tab:cond}. Again, finetuning lowers overall consistency, but results in less drop-off in non-adjacent vs. adjacent conditions.

\begin{table}[t]
    \centering
    \begin{tabular}{l|cc||cc}
\hline
    & \multicolumn{2}{c||}{Word 1 = US} & \multicolumn{2}{c}{Word 1 = UK}\\
    \hline

    \hline Condition & T5 & T5+FT & T5 & T5+FT\\
    \hline
 Adjacent &  92.2 & 71.1 &  65.1 & 63.4\\
 Non-adjacent & 88.7 & 77.7 & 54.3 & 63.2\\    \hline
%
%
    \end{tabular}
    \caption{Percent of test set examples for which each model prefers consistent over inconsistent spelling.}
    \label{tab:acc}
\end{table}



\subsection{Measure 3: mutual information} \label{sec:llr}

We also calculated the average mutual information (MI) across all prompt/target pairs in order to measure the strength of association between spelling conventions in both words. For each pair, we calculate four joint probabilities --- P(US,US), P(US,UK), P(UK,US), p(UK,UK). We assume these four probabilities make up the entire universe with respect to a particular prompt/target pair, and normalize them so they sum to 1. This also allows us to easily calculate marginal probabilities simply by adding the appropriate joint probabilities -- e.g., P(US prompt) = P(US,US) + P(US,UK). To calculate MI, 
we use a formula based on the log-likelihood ratio calculation in \citet{moore-2004-log}, but equivalent to the standard formulation for mutual information, where $x,y$ are the two probe words:
\[
\sum_{x \in \{\textsc{uk},\textsc{us}\} , y \in \{\textsc{uk},\textsc{us}\}} {p(x, y) \log \frac{p(x, y)}{ p(x)p(y)}} 
\]

Since T5 is trained on masked token prediction,
to measure the joint probability
$p(x,y)$ of each pair of probe words $x,y$ 
we can simply mask both probing tokens and measure the probability of generating both of them. 
That is, we present T5 with \ref{ex:t5in} and measure the probability of \ref{ex:t5out}:

\ex. \label{ex:t5} \a. \hspace*{-0.1in}My preferred words are <{\sc blank-}\\\hspace*{-0.1in}{\sc span}-1>, <{\sc blank-span}-2>, and tree.\label{ex:t5in}
 \b. \hspace*{-0.1in}<{\sc input-span}-1> flavour <{\sc input-}\\\hspace*{-0.1in}{\sc span}-2> harbour <{\sc input-span}-3> \label{ex:t5out}




\begin{table}[t]
    \centering
    \begin{tabular}{lcc}
\hline
    Condition & T5 & T5+FT \\
    \hline
 Adjacent & 0.0048 & 0.0017 \\
 Non-adjacent & 0.0044  & 0.0015  \\    \hline

    \end{tabular}
    \caption{Average mutual information in the adjacent and non-adjacent conditions.}
    \label{tab:llr}
\end{table}

Table \ref{tab:llr} presents these mutual information values. There doesn't seem to be a significant difference between adjacent and non-adjacent conditions for either T5 variant, though finetuning does seem to cause an overall drop in MI, in line with the overall drop in consistency seen in the measures above.

\subsection{Nonce forms}

We want to determine if T5 assigns the probabilities reported above on the basis of dependencies between specific lexical items, or if it is learning sub-word generalizations.
In other words, does the model learn that specific words like \textit{flavour} and \textit{realise} are more likely to co-occur than \textit{flavour} and \textit{realize}?
Or does it learn that words containing \textit{-our} are more likely to co-occur with words containing \textit{-ise}?
Since the model is trained using SentencePiece tokenization \cite{kudo-richardson-2018-sentencepiece}, it is possible that it exploits sub-word features.

One way of testing if a model can use sub-word features is to create nonce words that contain British- or American-specific sub-words.
If the model treats these as being British or American, this is an indication that the model is able to pick up on sub-word features.

We created a list of ten nonce forms by changing, adding, or removing one to three letters in existing words in our dictionary of American and British forms. 
These forms are shown in Table \ref{tab:nonceforms}.

\begin{table}[t]
    \centering
    \begin{tabular}{cc||cc}
\hline
British & American & British & American\\
\hline
glavour & glavor & reptalise & reptalize \\
mentre & menter & amolirise & amolirize \\
unulise & unulize & sphectre & sphecter \\
malvour & malvor & imminise & imminize \\
larbour & larbor & voitre & voiter \\
\hline

    \end{tabular}
    \caption{Nonce forms created by making one to three changes to words in the American-British dictionary.}
    \label{tab:nonceforms}
\end{table}

We use the same probing template and method as described above.
For each probe, we use a real American or British word for the first probe word, and one of the nonce forms shown in Table \ref{tab:nonceforms} for the second.
For this experiment we queried  the base T5 model in the adjacent condition.
The resulting conditional probability table is shown in Table \ref{tab:nonceresults}.

\begin{table}[t]
    \centering
    \begin{tabular}{llcc}
    \hline
 
    &&\multicolumn{2}{c}{Word 2} \\
   & &  US & UK \\
    \hline
    \multirow{2}{*}{Word 1} &  US & 0.68 & 0.32  \\
    & UK & 0.56 & 0.44 \\
\hline
    \end{tabular}
    \caption{Conditional probability table for nonce forms given by T5. The table shows the conditional probability of Word 2 (which is a nonce form), given Word 1. For each instance, the probability has been normalized over each condition (i.e., each row in the table).}
    \label{tab:nonceresults}
\end{table}

Table \ref{tab:nonceresults} shows that the patterns shown in Section \ref{sec:cond} above do not generalize very strongly to nonce forms.
The probabilities assigned to US forms following UK forms are on average higher than UK forms following UK forms.  However, the difference between these alternatives is attenuated compared to when Word 1 is a US form, indicating that (a) there is a heavy skew towards US spelling conditions in the training data; but (b) some sensitivity to the UK context, if not enough to counteract the high US form priors.  This suggests that the results in Table \ref{tab:cond} are to a large extent driven by lexical dependencies rather than any lower-level spelling patterns encoded by wordpieces.


%% file: gpt2.tex
\subsection{Autoregressive LLMs} \label{sec:gpt2}

\begin{table*}[t]
    \centering
    \begin{tabular}{l|c|cc||cc||cc}
    \hline
    && \multicolumn{2}{c||}{GPT2 (tgt only)} & \multicolumn{2}{c||}{GPT2 (to EOS)} & \multicolumn{2}{c}{OWT}\\
    Condition & Word 1 & \multicolumn{2}{c||}{Word 2} & \multicolumn{2}{c||}{Word 2} & \multicolumn{2}{c}{Word 2}\\ 
    & & US & UK & US & UK & US & UK\\
    \hline
    Adjacent &
     US & 0.87 & 0.13 & 0.69 & 0.31 & 0.95 & 0.05\\
    & UK & 0.36 & 0.64 & 0.51 & 0.49 & 0.34 & 0.66
 \\\hline
         Non-adjacent & 
     US & 0.83 & 0.17 & 0.66 & 0.33 & 0.95 & 0.05\\
    & UK & 0.49 & 0.51 & 0.54 & 0.46 & 0.28 & 0.72\\\hline
    \end{tabular}
    \caption{\footnotesize Conditional probability of Word 2, given a template with Word 1, given by GPT2 scored until the end of the target word only (tgt only) and scored until the end of the sentence (to EOS).  We also present the conditional probabilities from pairs found in the training corpus, OWT.}
    \label{tab:condgpt2}
\end{table*}

\begin{table}[t]
    \centering
    \begin{tabular}{l|cc||cc}
\hline
    & \multicolumn{2}{c||}{Word 1 = US} & \multicolumn{2}{c}{Word 1 = UK}\\
    \hline

    \hline Condition & GPT2 & GPT2 & GPT2 & GPT2\\
     & target & EOS & target & EOS\\
    \hline
 Adjacent &  94.2 & 70.1 &  70.8 & 49.4\\
 Non-adjacent & 92.5 & 67.5 & 54.6 & 45.1\\    \hline
%
%
    \end{tabular}
    \caption{Percent of test set examples for which each GTP2 scoring variant prefers consistent over inconsistent spelling.}
    \label{tab:accgpt2}
\end{table}

Many commonly-used LLMs (including T5) are trained to predict words in the input that have been masked out. Another common class of LLMs, however, are trained to perform next-word prediction instead. To examine how such autoregressive architectures handle spelling consistency, we experiment with OpenAI's GPT2 \cite{radford2019language}, which has a readily available open-source implementation through HuggingFace.\footnote{\url{https://huggingface.co/gpt2}}

As GPT2 is purely autoregressive, we cannot compute the probability that a particular probe word will fill a masked sentence span as easily as we could with T5. We can only efficiently compute the probability of a suffix given a prefix. Given this caveat, we have at least two options for assigning conditional probability scores, neither of which should be treated as exactly comparable to the T5 scores above. First, we can count only the logits corresponding to the target word:

P(``harbour'' | ``My preferred words are flavour,''). 

\noindent This local score ignores any words in the template occurring after the target word. Second, we can compute from the start of the target to the end of the sentence: P(``harbour, and tree'' | ``My preferred words are flavour,''), which accounts for the post-target suffix of the sentence.

Tables~\ref{tab:condgpt2} and~\ref{tab:accgpt2} show results for both of these methods for calculating the conditional probability, compiled in the same way as the T5 results in Tables~\ref{tab:cond} and~\ref{tab:acc}.  Table~\ref{tab:condgpt2} also includes the conditional probabilities from GPT2's training corpus, OWT. We see that GPT2 shows a similar preference for consistency as T5, but only very locally. There is a large drop-off in  preference for consistency when moving from adjacent to non-adjacent conditions, or when including the completion of the sentence in the calculation. For UK English in particular, any preference for consistency completely disappears beyond the immediate vicinity of the priming word, and the model returns to chance performance on the task.

%% file: exp2corpus.tex
\section{Further analysis of corpora}  \label{sec:corpuspartdeux}
We now return to a slightly more detailed examination of two of the corpora presented in Table \ref{tab:corpora}, English Wikipedia and the British National Corpus, both of which had relatively high levels of mismatch compared to the other corpora.

Wikipedia is an interesting case, since the documents are collectively edited by potentially a large number of contributors, which may lead to  higher expected mismatch than in other corpora.  For example, one version of the article on {\it air lock\/} used both US-spelling of the word {\it vapor\/} and the UK-spelling ({\it vapour\/}).  This is explained via three versions of the introductory sentence to the page, shown in Table \ref{tab:vapour} in Appendix \ref{sec:dataexamples}, where the two spellings are added to the sentence at different times, years apart.

The amount of mismatch in the British National Corpus is perhaps more surprising, given the provenance of the materials and intent of the collection. However the diversity of sources, which include things such as journal articles and edited volumes, likely leads to similar issues to those found in Wikipedia, along with simple human error and/or inconsistency.  Table \ref{tab:BNCexamples} in Appendix \ref{sec:dataexamples} presents a few examples of sentences with words from both spelling conventions, with American {\it -ize\/} spellings mixed with British {\it -ise\/} or {\it -our\/} versions.

%% file: conclusion1.tex
\section{Conclusion and Future Work}

We have presented results showing that T5 does tend towards consistency in spelling, but not to the degree that could be relied upon should such consistency be desired in generated text.
We show that this general preference for consistency reflects the data that the model is trained on, which also is mostly consistent, but with a significant proportion of exceptions.
The model's behavior is also shown to be affected by the relative frequency of language varieties in the training data.  We took advantage of the explicit and surface-accessible nature of these dependencies to attribute some model performance to the training data, while also demonstrating that modeling improvements should be possible, since the training data itself is substantially more consistent than the models.

These results suggest several possible avenues for future work.  First, methods for addressing bias in training data should yield improvements for British spelling consistency in these models.
We also intend to extend these results to languages other than English and investigate how spelling variation in other language situations is learned by LLMs.
Some of the methods we used here rely on the fact that English spelling variation is quite thoroughly catalogued.
Extending this work to less-documented cases of language variation will require us to either (1) collect data about spelling variation from language informants or data, or (2) develop methods that require less prior knowledge.
In the interest of finding methods that are extensible to the greatest number of cases, we intend to pursue path (2), working on methods to mine information about language variation from large corpora and LLMs that have been trained on them.

%% file: ack.tex
\section*{Acknowledgements} \label{sec:ack}
Thanks to Alexander Gutkin, Shankar Kumar, Arya McCarthy and Richard Sproat for useful discussion and comments, and to the anonymous reviewers for helpful suggestions.

%% file: limitations.tex
\section*{Limitations} \label{sec:limit}

Our work is focused on just a single case study of spelling variation. 
As detailed in Section \ref{sec:background}, English is a good candidate for a case study for several reasons, but it would be beneficial to extend this work to other language situations.

Another limitation was our choice to focus on already existing pre-trained models, rather than directly controlling the training data that is input to each model.
This means some of the conclusions about the connection between training data and outcome are tentative, pending experimental confirmation. 

%% file: ethics.tex
\section*{Ethics Statement}

This work does not propose a new model or dataset, but rather probes the behavior of existing models.
Thus novel ethical considerations about model behavior and dataset contents are not directly raised by this work.
While not explicitly focused on ethical considerations, this paper's methods hopefully contribute to better understanding model behavior, and could be used to understand the ways in which large language models treat underrepresented and marginalized language varieties.

%% file: appendix.tex
\section{Prompts}\label{sec:appendixprompt}
Table \ref{tab:PromptSet} presents the 29 prompt templates that were used in this study. Table~\ref{tab:stdvar} gives macro-averaged conditional probabilities for T5 runs across the different prompts, along with standard deviations to indicate how much performance varies due to the choice of prompt.

\begin{table*}[t]
\centering
\begin{tabular}{l}
\hline\hline
My preferred words are <CUE> and <FILLER>.\\
My preferred words are <CUE>, <FILLER>, and tree.\\
She wrote the words <CUE> and <FILLER>.\\
She wrote the words <CUE> and <FILLER> in her notebook.\\
She wrote the words <CUE>, <FILLER>, and cabbage.\\
I wrote the words <CUE> and <FILLER>.\\
I wrote the words <CUE> and <FILLER> in my notebook.\\
I wrote the words <CUE>, <FILLER>, and cabbage.\\
He wrote the words <CUE> and <FILLER>.\\
He wrote the words <CUE> and <FILLER> in his notebook.\\
He wrote the words <CUE>, <FILLER>, and cabbage.\\
We wrote the words <CUE> and <FILLER>.\\
We wrote the words <CUE> and <FILLER> in our notebook.\\
We wrote the words <CUE>, <FILLER>, and cabbage.\\
Mary wrote the words <CUE> and <FILLER>.\\
Mary wrote the words <CUE> and <FILLER> in her notebook.\\
Mary wrote the words <CUE>, <FILLER>, and cabbage.\\
Please spell <CUE> and <FILLER>.\\
Please spell <CUE>, <FILLER>, and panther.\\
Please spell <CUE> and <FILLER> correctly.\\
Say <CUE> and <FILLER>.\\
Say <CUE>, <FILLER>, and tapestry.\\
Say <CUE> and <FILLER> again.\\
The first words on the list were <CUE> and <FILLER>.\\
The first words on the list were <CUE>, <FILLER>, and oligarchy.\\
The easiest words on the list were <CUE> and <FILLER>.\\
The easiest words on the list were <CUE>, <FILLER>, and oligarchy.\\
The hardest words on the list were <CUE> and <FILLER>.\\
The hardest words on the list were <CUE>, <FILLER>, and oligarchy.\\
\hline\hline
\end{tabular}
\caption{Prompts used for model evaluation. Non-adjacent versions of each prompt were created by inserting the sequence ``, flower, interesting, jump, ponderous, sky, skipping, desk, small, ladder, lovely,'' between the <CUE> and <FILLER> word slots.}\label{tab:PromptSet}
\end{table*}

\begin{table*}[t]
    \centering
    \begin{tabular}{l|c|cc||cc}
    \hline
    && \multicolumn{2}{c||}{T5} & \multicolumn{2}{c}{T5+FT} \\
    Condition & Word 1 & \multicolumn{2}{c||}{Word 2} & \multicolumn{2}{c}{Word 2}\\
    & & US & UK & US & UK \\
    \hline
    Adjacent &
     US & 0.86 (0.01) & 0.14 (0.01) & 0.66 (0.03) & 0.34 (0.03) \\
    & UK & 0.39 (0.06) & 0.61 (0.06) & 0.44 (0.03) & 0.56 (0.03) 
 \\\hline
         Non-adjacent & 
     US & 0.83 (0.02) & 0.17 (0.02) & 0.69 (0.02) & 0.31 (0.02) \\
    & UK & 0.48 (0.05) & 0.52 (0.05) & 0.43 (0.04) & 0.57 (0.04) \\    \hline
    \end{tabular}
    \caption{Conditional probability of Word 2, given a template with Word 1, given by base T5 and T5 with additional finetuning. Each cell includes a macro-average and standard deviation across 29 prompts.}
    \label{tab:stdvar}
\end{table*}

\section{Examples of corpus mismatch}\label{sec:dataexamples}
Tables \ref{tab:vapour} and \ref{tab:BNCexamples} present examples illustrating the mixture of American and British spelling in Wikipedia and the British National Corpus, respectively, as discussed in Section \ref{sec:corpuspartdeux}.

\begin{table}[t]
\begin{tabular}{@{}p{0.8in}@{~}|p{2.1in}@{}}
version date & sentence version\\\hline
\href{https://en.wikipedia.org/w/index.php?title=Air_lock&oldid=793917943}{4 Aug. 2017} (neither vapor$\mathrm{~nor}$ vapour) & An {\bf air lock} is a restriction of, or complete stoppage of liquid flow caused by gas trapped in a high point of a liquid-filled pipe system.\\\hline
\href{https://en.wikipedia.org/w/index.php?title=Air_lock&oldid=799204679}{6 Sept. 2017} (vapour replaces gas) & An {\bf air lock} is a restriction of, or complete stoppage of liquid flow caused by vapour trapped in a high point of a liquid-filled pipe system.\\\hline
\href{https://en.wikipedia.org/w/index.php?title=Air_lock&oldid=940980759}{15 Feb. 2020} ~~~(vapor added) & An {\bf air lock} (or {\bf vapor lock}) is a restriction of, or complete stoppage of liquid flow caused by vapour trapped in a high point of a liquid-filled pipe system.\\\hline
\end{tabular}
\caption{Three versions of a Wikipedia page: (1) no use of {\it vapor\/} or {\it vapour\/} in the sentence; (2) the term {\it vapour\/} replaces {\it gas\/}; and (3) the alternative name for the phenomenon "{\it vapor lock\/}" is introduced.}\label{tab:vapour}
\end{table}

\begin{table}[t]
\begin{tabular}{@{}l@{~}|p{2.4in}@{}}
Doc ID & sentence\\\hline
CPD & `What this guy will do is get a {\bf demoralized} sales {\bf organisation revitalised}...' said John Jones, analyst at Salomon Brothers.
\\\hline
CLW & They {\bf conceptualize} these differences in terms of `separate local {\bf labour} market cultures' (ibid., p. 104).\\\hline
CBH & It is a metaphor which attempts to create a reality of {\bf organization} whereby cooperation is {\bf mobilised} for fight with the outside world.\\\hline
\end{tabular}
\caption{Examples of spelling convention mismatches in the British National Corpus, sampled from varied books and periodicals.}\label{tab:BNCexamples}
\end{table}